\documentclass[conference]{IEEEtran}

\makeatletter

\def\ps@IEEEtitlepagestyle{%
  \def\@oddfoot{\mycopyrightnotice}%
  \def\@evenfoot{}%
}
\def\mycopyrightnotice{%
  {\footnotesize 978-1-6654-6159-7/22/\$31.00~\copyright~2022 European Union\hfill}
  \gdef\mycopyrightnotice{}
}

\IEEEoverridecommandlockouts
\usepackage{cite}
\usepackage{amsmath,amssymb,amsfonts}
\usepackage{algorithmic}
\usepackage{graphicx}
\usepackage{textcomp}
\usepackage{xcolor}
\usepackage{booktabs}
\usepackage{todonotes}
\usepackage[english]{babel}
\usepackage{lipsum}
\usepackage{float}
\usepackage{multirow}
\usepackage{graphicx}

\def\BibTeX{{\rm B\kern-.05em{\sc i\kern-.025em b}\kern-.08em
    T\kern-.1667em\lower.7ex\hbox{E}\kern-.125emX}}

\graphicspath{{./images/}}

\usepackage{blindtext}
\usepackage{eso-pic}

\newcommand\AtPageUpperMyright[1]{\AtPageUpperLeft{%
 \put(\LenToUnit{0.17\paperwidth},\LenToUnit{-2cm}){%
     \parbox{0.9\textwidth}{\raggedleft\fontsize{8}{11}\selectfont #1}}%
 }}%
\newcommand{\conf}[1]{%
\AddToShipoutPictureBG*{%
\AtPageUpperMyright{#1}
}
}

\begin{document}
\title{\vspace*{1cm}Mapping Researcher Activity based on Publication Data by means of Transformers}

\author{\IEEEauthorblockN{Zineddine Bettouche and Andreas Fischer}
\IEEEauthorblockA{Deggendorf Institute of Technology \\
Dieter-Görlitz-Platz 1\\
94469 Deggendorf \\
E-Mail: zineddine.bettouche@th-deg.de, andreas.fischer@th-deg.de}

}

\maketitle

\conf{\textit{  Proc. of the Interdisciplinary Conference on Mechanics, Computers and Electrics (ICMECE 2022)  \\ 
6-7 October 2022, Barcelona, Spain}}

\begin{abstract}
Modern performance on several natural language processing (NLP) tasks has been enhanced thanks to the Transformer-based pre-trained language model BERT. We employ this concept to investigate a local publication database. Research papers are encoded and clustered to form a landscape view of the scientific topics, in which research is active. Authors working on similar topics can be identified by calculating the similarity between their papers. Based on this, we define a similarity metric between authors.
Additionally we introduce the concept of self-similarity to indicate the topical variety of authors.
\end{abstract}



\section{Introduction}
One of the central themes in natural language processing (NLP) is the task of text representation, which is a kind of rule for converting natural language input information into machine-readable data. Today, the most advanced text models use Transformers to teach how to represent text. Transformers are a
type of neural network that are increasingly finding their use in various branches of machine learning, most often in
sequence transduction problems, that is, such problems when both the input and output information is a sequence.

BERT (Bidirectional Encoder Representations from Transformers)~\cite{bert} has stirred up the machine learning world since its publication by showcasing cutting-edge results in a wide range of NLP tasks. The main technical advancement of BERT is the combination of left-to-right and right-to-left training with Transformer's bidirectional training (attention model) for language modeling. The study's findings demonstrate that bidirectionally trained language models can comprehend context and flow of language more deeply than single-direction language models. The data could be clustered more effectively by using the embeddings-format produced from running textual data via BERT as opposed to more conventional clustering techniques like topic extraction and text similarity.

Therefore, we attempt in this work to cluster a dataset of research papers (institute's internal database) into topics using Transformer models. The obtained clusters would provide an overview of the published-research landscape. Having met an acceptable clustering efficiency, authors working in similar topics could be linked together, based on the similarity between their papers. 

As for the structure of this paper, section~\ref{background} introduces the technologies used in this paper as a general background. Section~\ref{related-work} discusses the previous papers that dealt with document similarity and clustering, BERT, and keyword extraction. Section~\ref{data-analysis} provides an analysis on the data used in this work. Section~\ref{methodology} discusses the methodology and the overall implementation of the processes implemented for this paper. Section~\ref{experiments} presents the experiments done in this work, their implementation, and the rationale behind them. Finally, section~\ref{conclusion} concludes the work and sets up the possible future developments, which could be built on our findings, or, in the least, complement it in certain areas.   

\section{Background}
\label{background}

In this section, the main techniques used in this paper are discussed. 

\subsection{Transformer Models}
A deep learning technique for natural language processing (NLP) called Bidirectional Encoder Representations from Transformers (BERT) aids artificial intelligence (AI) programs in comprehending the context of ambiguous words in text.

By simultaneously evaluating text in both the left-to-right and right-to-left directions, BERT-using applications are able to forecast the accurate meaning of a synonym. Deep learning neural networks may use unsupervised learning methods to develop new NLP models thanks to BERT's bidirectionality, a masking strategy, and learning how to anticipate the meaning of an ambiguous term. This method of natural language understanding (NLU) is so effective that Google advises customers to use it to train a cutting-edge question and answer system in a short amount of time as long as there is sufficient training data available.

\subsection{UMAP: Dimensionality Reduction}
Similar to t-SNE, the dimensionality-reduction method known as Uniform Manifold Approximation and Projection (UMAP)~\cite{umap} can be utilized for visualization as well as generic non-linear dimensionality reduction. The following suppositions about the data form the basis of the algorithm:
\begin{itemize}
	\item The Riemannian metric is roughly constant locally, and the data are uniformly distributed on the Riemannian manifold.
	\item The manifold is connected locally.
\end{itemize}
These presumptions allow one to construct a fuzzy topological model of the manifold. Finding the embedding involves looking for a low-dimensional projection of the data that has the most similar fuzzy topological structure to the original data.

\subsection{K-means Clustering}
One of the most straightforward and well-liked unsupervised machine learning methods is K-means clustering~\cite{kmeans}. Unsupervised algorithms typically draw conclusions from datasets using only input vectors without taking into account predetermined or labelled results. 

The K-means algorithm finds k centroids, keeps the centroids as small as feasible, and then assigns each data point to the closest cluster. Finding the centroid is what `means' in the K-means algorithm indicates: averaging the data. The K-means technique in data mining uses a first set of centroids that are randomly chosen as the starting points for each cluster to process the learning data. 

The program then performs iterative (repetitive) calculations to optimize the positions of the centroids. It ends creating and optimizing clusters when either the centroids have stabilized, or the defined number of iterations has been achieved.

\section{Related Work}
\label{related-work}
Concerning related work, Beltagy et al. have published SciBERT~\cite{Beltagy2019}, a pretrained language model for scientific text based on BERT, in a paper titled `SciBERT: A Pretrained Language Model for Scientific Text'. They evaluated SciBERT on a suite of tasks and datasets from scientific domains. SciBERT significantly outperformed BERT-Base and achieves new SOTA results on several of these tasks, even compared to some reported BIOBERT results on biomedical tasks.

In a paper~\cite{ostendorff} titled `Aspect-based Document Similarity for Research Papers' published on a related subject, Ostendorff et al. apply pairwise multi-label multi-class document classification to scientific papers in order to determine an aspect-based document similarity score. Based on the paper's title and abstract, the investigated models are trained to forecast citations and the recognised label. Over two scientific corpora, they assess the Transformer models BERT, Covid-BERT, SciBERT, ELECTRA, RoBERTa, and XLNet along with an LSTM baseline. Overall, SciBERT outperformed all other models in the tests.
SciBERT predicted the aspect-based document similarity with F1-scores of up to 0.83 despite the difficult assignment. Transformers are highly adapted to accurately compute the aspect-based document similarity for research papers, according to their empirical investigation.

Chandrasekaran and Mago conducted a survey~\cite{Chandrasekaran2022} titled `Evolution of Semantic Similarity - A Survey' in which they stated that, measuring semantic similarity between two text snippets has been one of the most challenging
tasks in the field of Natural Language Processing. They concluded that, most recent hybrid methods
have shown promising results over other independent models. While the focus of recent research is
shifted towards building more semantically aware word embeddings, and the transformer models
have shown promising results, the need for determining a balance between computational efficiency
and performance is still a work in progress. 

The problem of semantic textual similarity in medical data was addressed by Kades et al. in a paper~\cite{kades} titled `Adapting Bidirectional Encoder Representations from Transformers (BERT) to Assess Clinical Semantic Textual Similarity: Algorithm Development and Validation Study.' The authors developed three methods to address this issue. They suggested enhancing BERT with new features and comparing several regression models based on the BERT result and other features. The use of M-Heads and an effort to automatically extrapolate medical knowledge from the training data was another concept. They noticed that the underlying dataset had a significant impact on the effectiveness of the various techniques.

`Measurement of Semantic Textual Similarity in Clinical Texts: Comparison of Transformer-Based Models' is an article published in the journal Clinical Text by Yang et al.~\cite{yang}, in which they showed how to measure clinical STS using transformer-based models, and created a system that can employ several transformer algorithms. In comparison to other transformer models, their experiment findings demonstrate that the RoBERTa model performed the best. The study also showed how well transformer-based models performed when used to evaluate the semantic similarity of clinical content.

In another paper~\cite{kretschmann} titled `Extracting Keywords from Publication Abstracts for an Automated Researcher Recommendation System' Kretschmann et al. presented a keyword assignment system based on an older version of the DIT publication database. It handles low volume data and missing keywords by extending the data volume using information from online publication databases, extending the total volume to 6500 items. A prototype keyword assignment system was built, that uses random oversampling in preprocessing and LightGBM as classifier with binary relevance as transformation method. As an enhancement of this system, Transformer models could be used to uncover relations between papers and authors on another level.

Therefore, our work investigates the use of Transformer models, mainly SciBERT, in the last mentioned recommendation system, as an attempt to enhance the quality of recommendations, through understanding the relations between abstracts deeply, using BERT models and K-means clustering.

\section{Exploratory Data Analysis}
\label{data-analysis}
The data used in this work are encoded in a JSON file with a size of 11.735 KB.
Contained are 7548 references of various types: the file includes
references to newspaper articles, blog entries, talks, along with scientific
articles such as conference papers or journal articles. Only the latter ones come
with abstracts, comprising a total of 1500 documents, which are chosen
as the data basis for this investigation.

\begin{figure}
    \centering
    \includegraphics[width=0.85\linewidth]{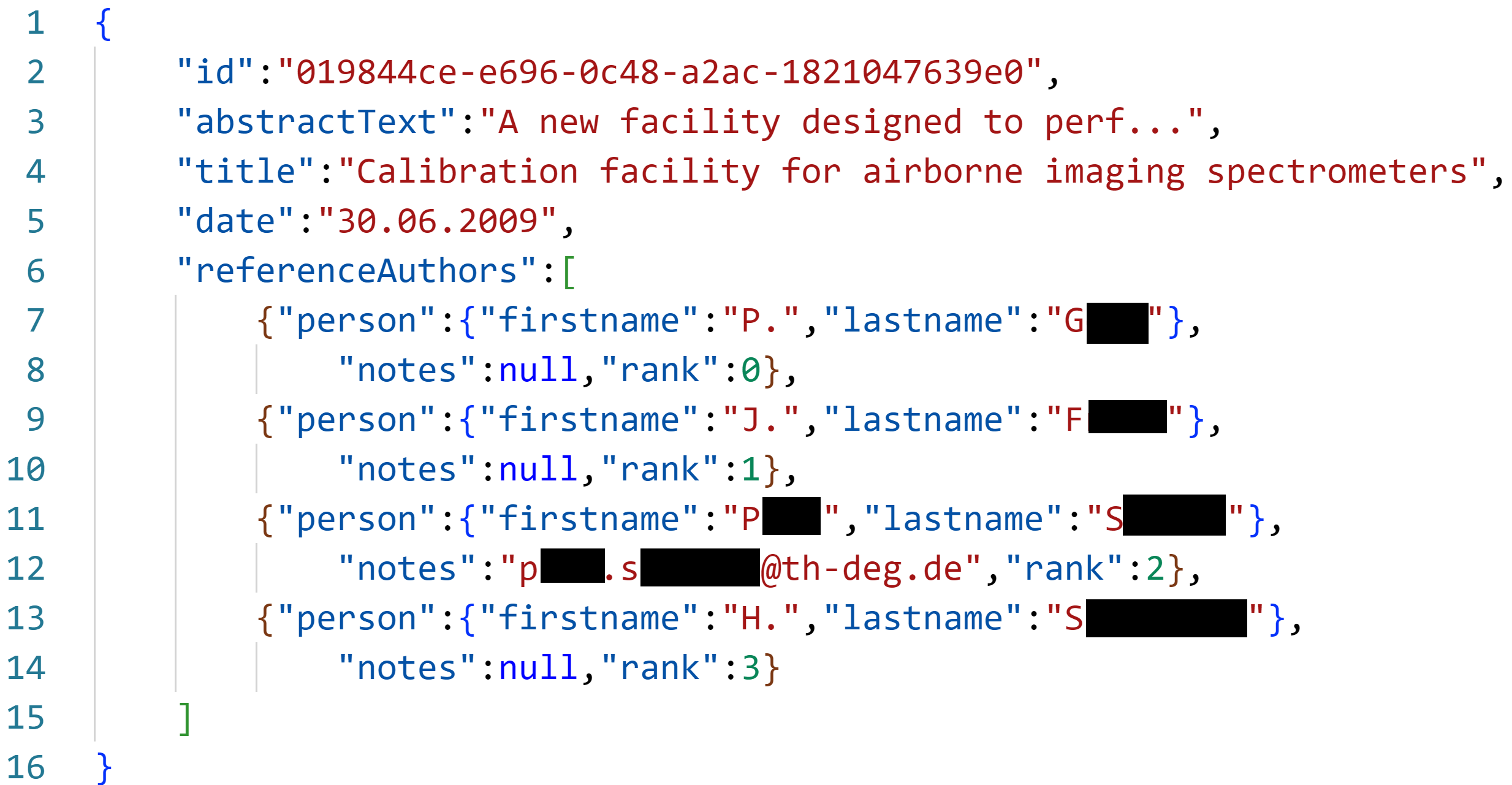}
	\caption{Paper-Object Example}    
	\label{fig:data-example}
\end{figure}

Each of the selected entries has at least a title, a list of authors, a date,
and an abstract. Figure~\ref{fig:data-example} shows an example of an entry. Each author is marked by
a name. It is possible to distinguish between internal authors (i.e., employees)
and external authors: Internal authors are identified by their e-mail address.
External authors, on the other hand, cannot reliably be distinguished. Two entries
with the same name as an external author might stem from the same person or
from two persons with the same name.

\begin{figure}
    \centering
    \includegraphics[width=0.98\linewidth]{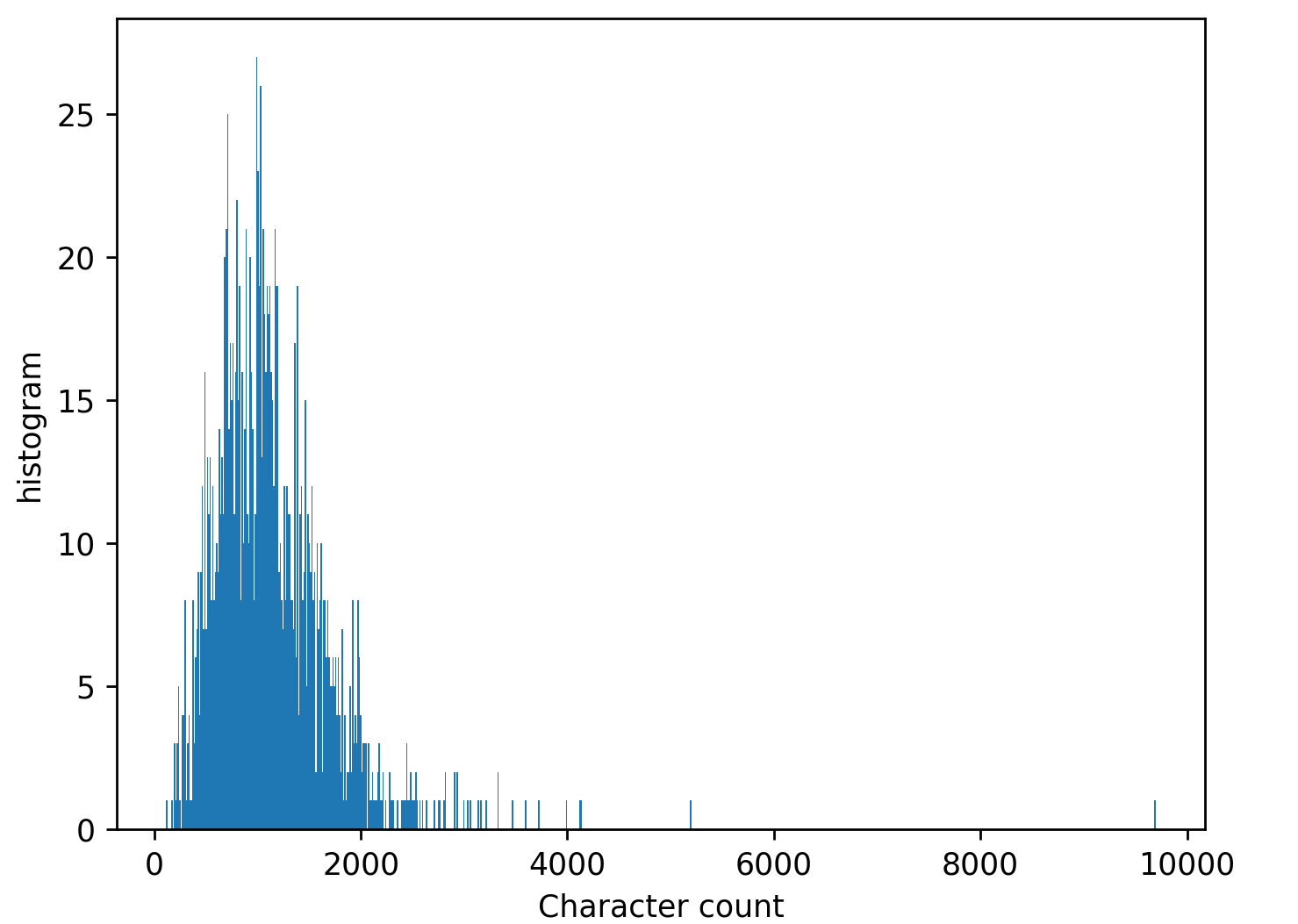}
	\caption{Histogram plot of the character count in each abstract}    
	\label{fig:histogram}
\end{figure}

The character count of every abstract present in the database was calculated, and the overall distribution is presented in Figure~\ref{fig:histogram}. It can be noticed that the distribution of data appears to represent a long-tail distribution, with the majority of documents falling in the interval of 500 to 1500 characters per abstract-text.
The entry at the far right is an outlier, in which the entire paper content is
recorded in the abstract field.

\section{Methodology}
\label{methodology}
In this section the general approach is described, followed by a discussion of the metric used to calculate author distances.

\subsection{General Approach of Processing}
\begin{figure}
    \centering
  \includegraphics[width=0.95\linewidth]{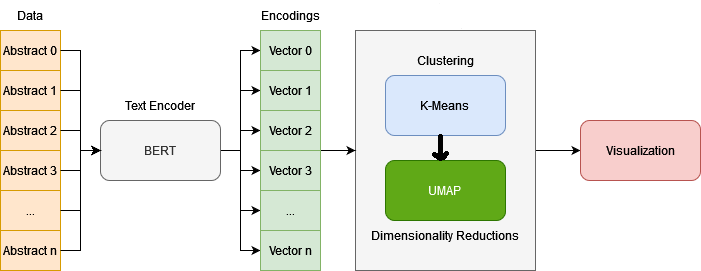}
  \caption{Implementation Overview}
	\label{fig:implementation-overview}
\end{figure}

The goal of the implementation is to cluster the encodings of the abstracts and visualize it, in order to assess the performance of BERT when dealing with relatively long textual data. The starting point of the implementation is getting the data as input. The data, as previously mentioned, is in the form of paper objects. Every object contains essentially an ID and an abstract text.  Therefore, the abstract-texts are encoded with a BERT encoder, which gives in return the vector format (768-dimension) for each text. The dimensions of the obtained vectors must be reduced into 2D in order for visualization to take place. Hence, K-means can be applied on the 768D vectors, and then the dimensionality-reduction is done for graphing. The implementation overview is displayed in Figure~\ref{fig:implementation-overview}.

In order to obtain similar papers, Euclidean distance is calculated between the encodings of each abstract-text. This attempt can display the efficiency of clustering textual data based on its BERT enconding, rather than traditional words-sequence assessment.

\subsection{Calculating distances between authors}

The embedding of documents produced by transformer architectures such as the
BERT family has, for the use case outlined in this paper, the advantage that
a distance between authors can be directly derived from document distances.
We define the distance between two authors as the mean pair-wise distance
of their respective papers. I.e., let $P_1$ be the set of papers by
author 1 and $P_2$ likewise be the set of papers by author 2. Then the distance
between authors 1 and 2 is defined as:

\begin{equation}
	\frac{\sum_{p_1 \in P_1} \sum_{p_2 \in P_2} \text{dist}(p_1, p_2)}{|P_1| \cdot |P_2|}
\end{equation}

There are a few border cases to be discussed here. First, there will be a distance
for every pair of authors, provided both of them have published a paper. If an
author is recorded in the database without any publication, no useful distance
can be computed to any other author. This is to be expected.

Second, papers may appear both in $P_1$ and $P_2$---in particular for co-authors.
In this case, the corresponding distance for that paper will, of course, be zero.
This lowers the overall distance, and is an expected effect.

Third, an author can be compared to itself. The overall
result would \emph{not} be zero (as could be expected), but instead report the average
distance among his papers. We argue that this is useful, as it provides
a measure of the self-similarity, i.e., the topical variety of the papers from a given author.

\section{Experiments}
\label{experiments}
In this section, the experiments done throughout this work and both the rationale behind them, and the results obtained from them, are discussed.

\subsection{SciBERT: From Abstracts To 768D-Vectors}
\begin{figure}
    \centering
    \includegraphics[width=0.9\linewidth]{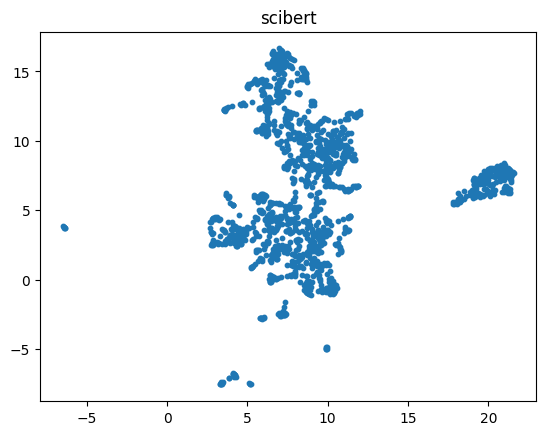}
	\caption{2D representations of 768D-encodings resulting from SciBERT encoding the abstracts}    
	\label{fig:scibert-scatter}
\end{figure}

As we have already mentioned in the background and related-work sections, SciBERT scores best when dealing with scientific papers. We attempt now to obtain vector-representations of each of the abstracts in the data, which are visualized after being reduced to 2D points by UMAP. Figure~\ref{fig:scibert-scatter} shows the resulting scatter graph.

An initial remark could state that, the points experience a few condensations. This indicates the presence of clusters in the data. Therefore, we proceed to cluster the data using K-means.

\subsection{Initial Raw Clustering: K-means Application}
Before we indulge in describing how the different techniques in this work (such as K-means and UMAP) are used together, it is worth to mention that the process of clustering is implemented in the our code, in a way that automates the discovery of clusters-number in the data, instead of assuming it. The program loops n times (n is in the range [10, 30]) and selects the number of clusters attached to the best silhouette score. The silhouette value~\cite{silhouettes} is a measure of how similar an object is to its own cluster (cohesion) compared to other clusters (separation).

After encoding the abstract-texts and obtaining the 768D-vectors, they can be clustered first, then mapped into 2D-vectors, which are then visualized. A hypothesis can state that clustering directly the 768D-vectors would give meaningful clusters, on the condition that the similarity between two abstract-texts is inversely proportional to the Euclidean distance between each of their vectors.
It is critical to understand that the compactness of clusters does not imply the efficiency the encodings of the abstract-texts. Figure~\ref{fig:clusters-768D-notes} shows the obtained clusters and the titles of some interesting cases, that we would like to comment on as follows:
\begin{figure}
	\centering
	\includegraphics[width=\linewidth]{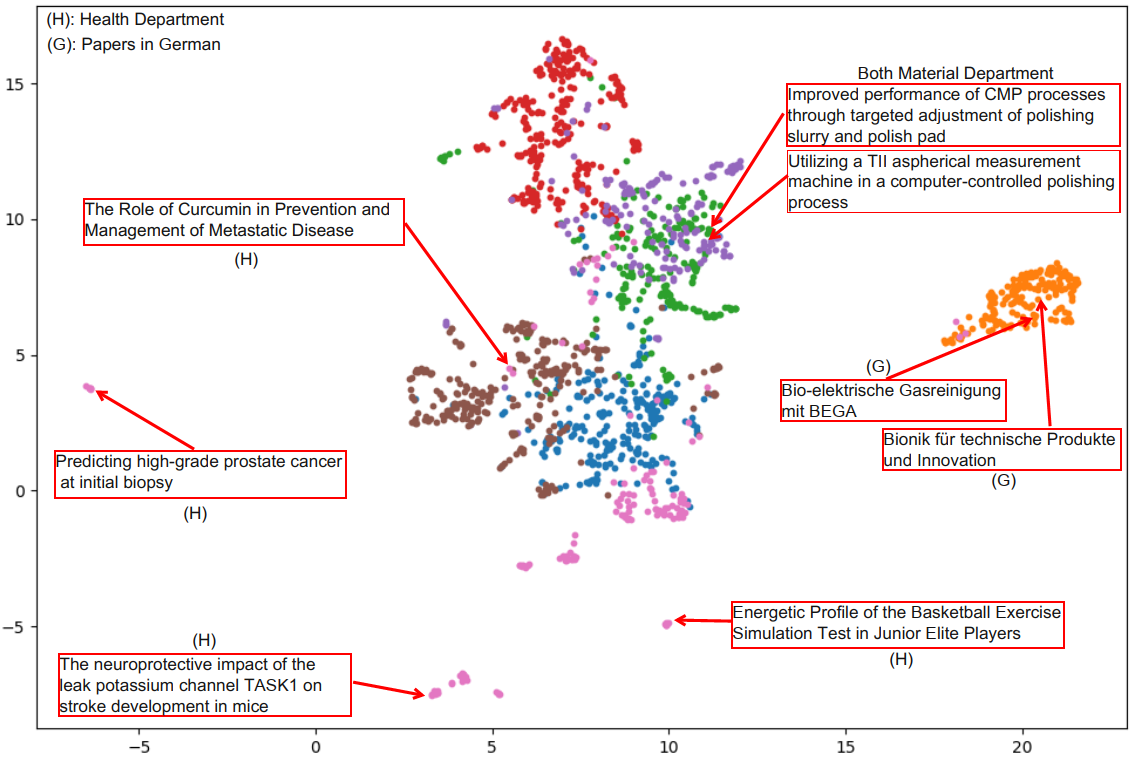}
	\caption{Detailed points (768D-vector clustering)}
	\label{fig:clusters-768D-notes}
\end{figure}
\begin{itemize}
	\item In the case of health-related papers (pink), although some points were relatively farther from the centroid of their cluster than most of that cluster's points, K-means was accurate enough to assign them properly. This indicates that, the inaccurate representation of the distance on the graph (i.e. when it does not indicate the actual similarity between two paper), is due to the nature of UMAP, when it attempts to map 768D-vectors into a 2D points. This is regarded as a positive result concerning the relevancy of the 768D-vectors to one another, when their original abstract-texts are similar.
	\item It is remarked from the distinct cluster on the right-hand side (orange) that, the encoder used (SciBERT) has separated papers written in German from papers written in English. This can be further investigated by extracting the German papers, and running the encoding process on them separately. 
	\item Concerning the other clusters, we can say that, both the distances between the points and their assigned clusters were reasonably accurate; as they were grouped in the form of: optics in red, networks in blue, and media in brown (including image/video processing).
	\item Finally, we have reason to believe that the automated silhouette method, which determines clusters-number for K-means, has made an unnecessary split, resulting in two clusters (purple/green) that belong to the same department (Material Department). Going further, clusters-number can be manually defined at this point, which will shed more light onto their global cluster, when rerunning K-means.
\end{itemize}
Therefore, in this paper, we would proceed by casting the German papers aside, calculating cluster-metrics, and finally extracting keywords for each obtained cluster.

\subsection{Clusters: Metrics and Keywords}
\begin{figure}
	\centering
	\includegraphics[width=\linewidth]{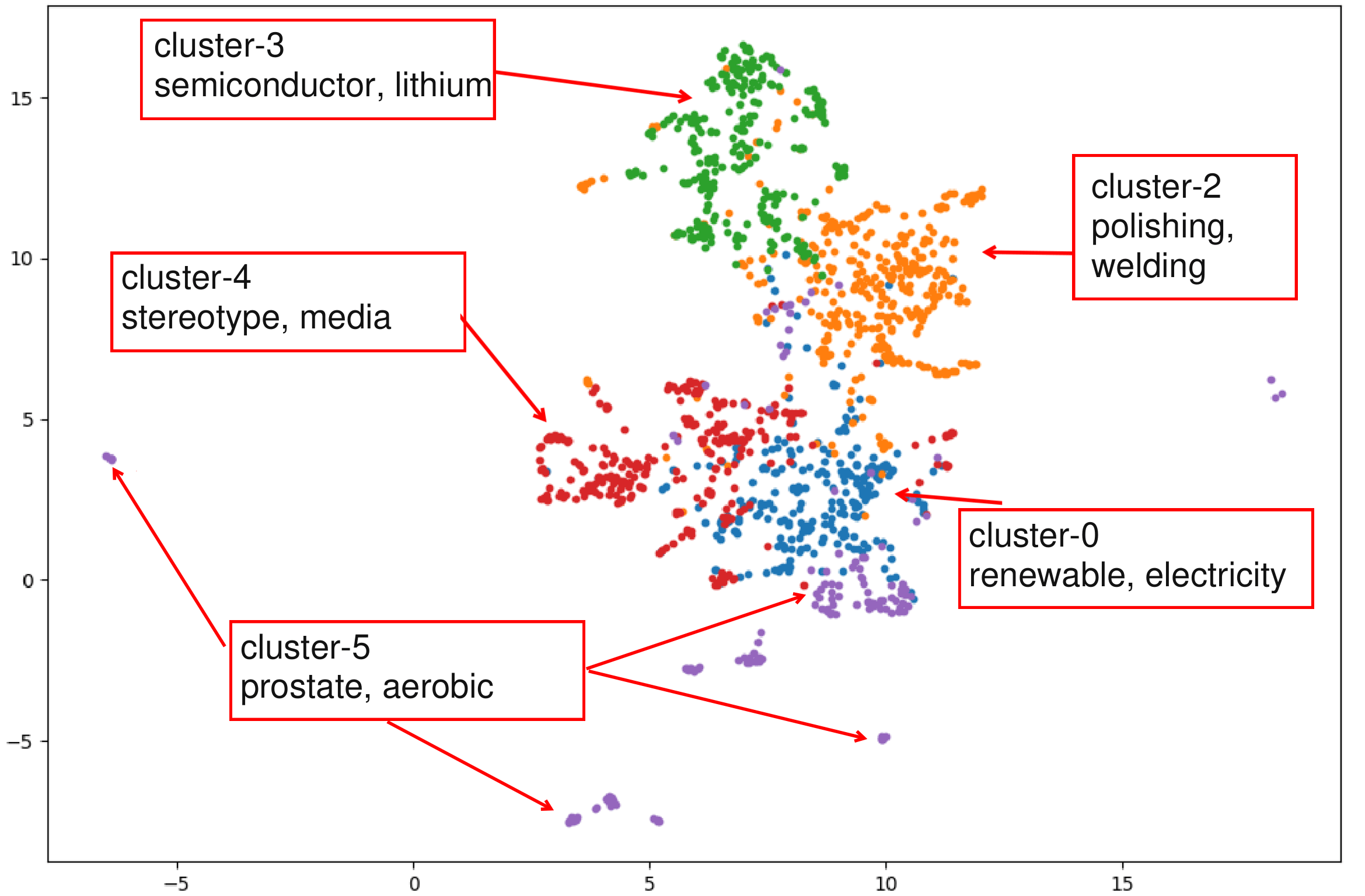}
	\caption{Adjusted clusters}
	\label{fig:updated-clusters}
\end{figure}

As a first step in extracting the keywords of each cluster, and potentially, the topic of each, the German data was separated from the rest of the data (mostly English). The latter was adjusted further, based on what we perceived as one cluster divided unnecessarily in half. Figure~\ref{fig:updated-clusters} shows the adjusted clusters and a few keywords of each cluster.

To finally assess the clusters with common metrics, some of the clustering metrics were calculated and are as follows:
\begin{itemize}
\item Silhouette: 0.103
\item Calinski-Harabasz: 129.237
\item Davies-Bouldin: 3.085
\end{itemize}

Keywords could be extracted using the KeyBERT library, we then attempt to assign a topic to such keywords accordingly. 
In addition, a further evaluation of the clusters can be set in terms of the radius and the standard-deviation of each cluster. Table~\ref{tab:metrics-table} shows the result of our calculations and keyword-extraction.

\begin{table*}
  \centering
\caption{Clusters-Metrics Table}
  \label{tab:metrics-table}
\begin{tabular}{p{1cm}p{1cm}p{1cm}p{1cm}p{5cm}p{3cm}}
 \toprule
\multicolumn{6}{c}{English Papers} \\
 \midrule
label&768D-radius&768D-std&points-count&keywords&topic\\
 \midrule
 0&4.903&0.708&251&renewable, photovoltaic, electricity&Power\\
1&3.846&1.080&178&german papers (ignored)&german papers (ignored)\\
2&5.048&0.830&369&polishing, welding, piezoelectric&Production/Industry\\
3&4.773&0.647&275&semiconductor, nanowire, lithium&Material\\
4&4.801&0.694&279&stereoscopic, 3dtv, multimedia&Media\\
5&5.140&0.832&148&prostate, aerobic, schizophrenia&Health\\
\bottomrule
\end{tabular}
\end{table*}

\subsection{Distances between Authors}
\begin{figure}
	\centering
	\includegraphics[width=\linewidth]{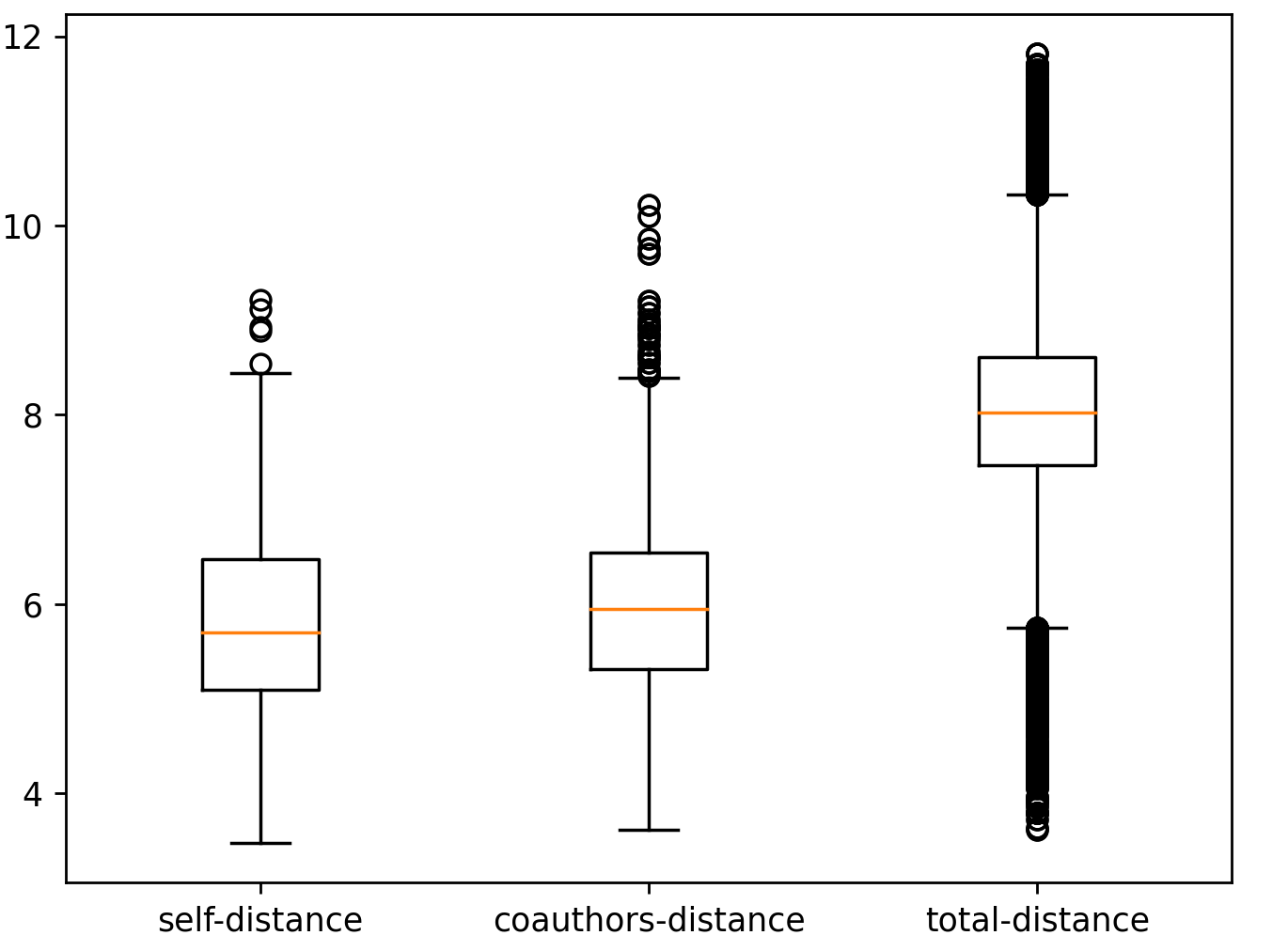}
	\caption{Author-distance distributions in the three mentioned cases}
	\label{fig:authors-distances}
\end{figure}

As explained before, how close two authors are, is related to how similar are their papers. Therefore, to evaluate whether the distances, obtained in this work between authors, are reasonable, we have investigated the difference between three cases: the average distance each other has to himself (self-distance), the average distance between co-authors (coauthor-distance), and the total average distance between all authors. Figure~\ref{fig:authors-distances} shows the author-distance distributions in the three mentioned cases.
It can be seen that, as could be expected, self-distance and coauthors-distance are relatively
close, with coauthors-distance slightly higher. On the other hand, there is a significant difference
to the average distance between all authors. Ergo, the defined distance metric for authors captures
semantic relationship and, as such, is a useful tool to indicate the degree of overlapping research
for two given authors.

\begin{figure}
	\centering
	\includegraphics[width=\linewidth]{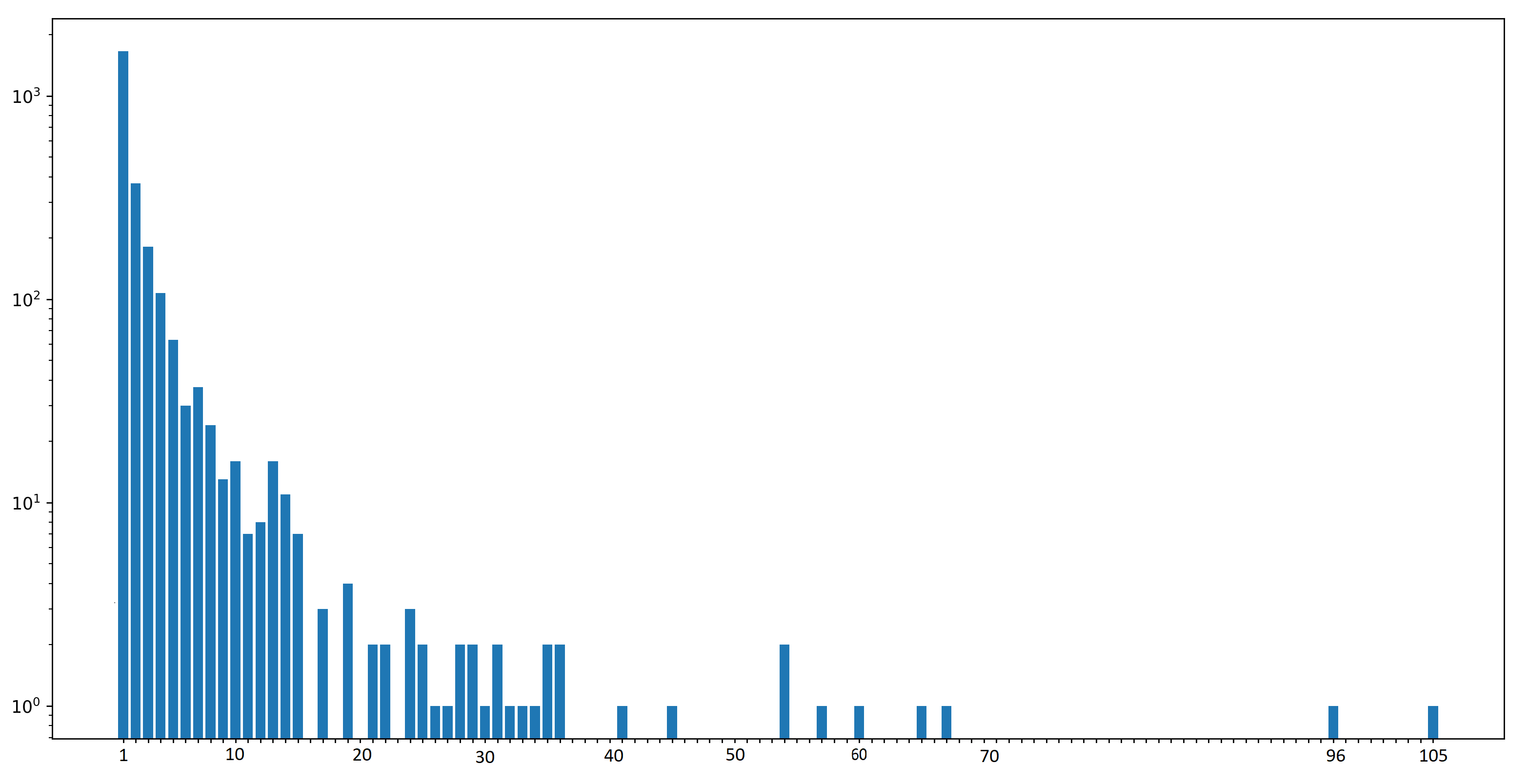}
	\caption{Distribution of paper count per author}
	\label{fig:papers-per-author}
\end{figure}

For further analysis, only active authors with ten publications or more are considered to avoid
bias that results from a low number of publications.
Figure~\ref{fig:papers-per-author} shows the distribution of paper count per author. As it can be seen, the majority of authors have 1 to 15 published papers, whereas the highest paper count is 106. Logarithmic scale was used on the y-axis, due to the huge difference between authors with number of papers lower and higher than 10. 

To further visualize the self-distance of authors, we excluded the authors having fewer than 10 published papers (100 out of 4426 authors), because having very few papers could actually distort the distance calculations. Figure~\ref{fig:papers-per-author} shows the distribution of paper count per author. 

\begin{figure}
	\centering
	\includegraphics[width=\linewidth]{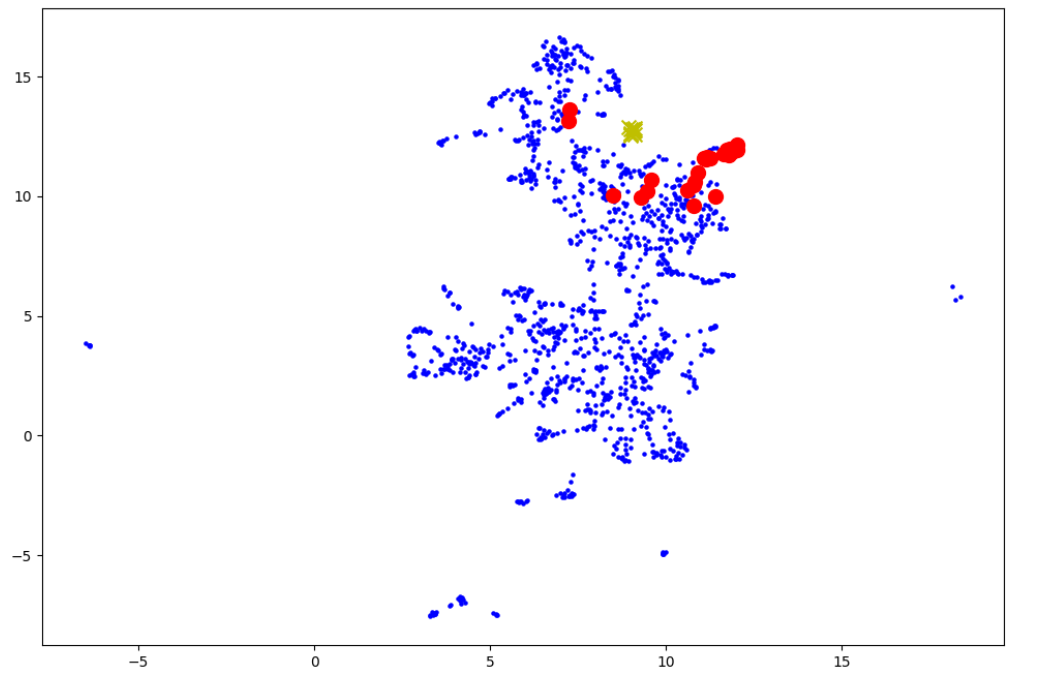}
	\caption{The papers of the two authors having the highest and lowest self-distances (red and yellow, respectively)}
	\label{fig:authors-cases}
\end{figure}

German-papers were again excluded to avoid confusion between topical clusters and language clusters.
The two authors with the highest vs.~the lowest self-similarity are selected and visualized in Figure~\ref{fig:authors-cases}.
The author represented by the red blobs (highest self-distance with 30 papers) is active in several
topical areas, such as production, industry, and materials engineering.
On the other hand, the author represented by the yellow crosses (lowest self-distance with 10 papers)
is focused on the more narrow field of Materials.

\section{Conclusion}
\label{conclusion}
In this paper, SciBERT encodings were used as a means to obtain vector representations of paper abstracts.
Semantic similarity between papers is encoded in the pairwise distance of vectors.
With the application of K-Means, topical research clusters were identified in a given publication data base.
Visualization using UMAP highlights the topical areas. By extracting keywords from the clustered papers,
the research areas could be identified. Based on distance between papers, a distance metric between
authors was introduced. The data indicates that this distance metric is a useful tool to indicate
topical relationships between authors.

For this paper, a cluster of purely German articles was ignored to avoid confusion between topical
and linguistical similarity. Future work includes developing an approach that can reliably handle
multilingual data. Further investigation is also needed in how to apply the author similarity
metric in a recommender system. Finally, more research is required on the keyword extraction mechanism
used for cluster labeling: The current approach is based on manual investigation of extracted keywords.
An ontology-based system might be able to automate this process and, such, scale to larger publication
databases.

\section*{Acknowledgement } 
This paper has received funding from the state of Bavaria in the context of project SEMIARID, funding no. DIK-2104-0067//DIK0299/01

\bibliographystyle{IEEEtran}
\bibliography{references}

\end{document}